\documentclass[runningheads]{llncs}
\usepackage[T1]{fontenc}
\usepackage{subcaption}
\usepackage{graphicx,verbatim}
\usepackage{listings}
\usepackage{minted}
\usepackage{fontspec}
\usepackage{booktabs}
\usepackage{array}    
\usepackage{tabularx} 
\usepackage{ragged2e} 
\usepackage{float}
\usepackage{siunitx}
\usepackage{amssymb}
\usepackage{enumitem}

\begin{document}

\title{LongMedBench: Benchmarking Medical Agents for Long-Horizon Clinical Decision-Making}
\titlerunning{LongMedBench}
%


\author{Zihan Xu\inst{1}\textsuperscript{*} \and
Yanzhen Chen\inst{1}\textsuperscript{*} \and
Xiaocheng Zhang\inst{1}\textsuperscript{*} \and
Zhiting Fan\inst{1} \and
Weiqi Zhai\inst{2} \and
Hongxia Xu\inst{1,3} \and
Zuozhu Liu\inst{1,3}\textsuperscript{\ensuremath{\dagger}}}
\index{Chen, Yanzhen}
\index{Xu, Zihan}
\index{Zhang, Xiaocheng}
\index{Fan, Zhiting}
\index{Zhai, Weiqi}
\index{Xu, Hongxia}
\index{Liu, Zuozhu}

\authorrunning{Z. Xu et al.}

\institute{Zhejiang University, Hangzhou, China\\
\email{\{zihan1.22,yanzhen.22,xiaocheng.22,zhiting.23\}@intl.zju.edu.cn}
\email{einstein@zju.edu.cn, zuozhuliu@intl.zju.edu.cn}
\and
Alibaba Group, Hangzhou, China\\
\email{zhaiweiqi.zwq@alibaba-inc.com}
\and 
Transvascular Implantation Devices Research Institute, Hangzhou, China
}
\maketitle
\begingroup
\renewcommand{\thefootnote}{}
\footnotetext[1]{\textsuperscript{*} These authors contributed equally to the work.}
\footnotetext[2]{\textsuperscript{\ensuremath{\dagger}} Corresponding author: Zuozhu Liu.}
\endgroup
\setcounter{footnote}{0}

\begin{abstract}
In this work, we introduce LongMedBench, a real-world EHR-based benchmark for long-horizon clinical decision-making. Prior evaluations of LLM-based medical agents have largely emphasized short-context knowledge QA and tool use. However, real-world medical care is inherently longitudinal, and clinicians must aggregate evidence across repeated visits, tests, and evolving treatments. Therefore, long-horizon interaction is essential for realistic assessment. LongMedBench is constructed via a reproducible pipeline that integrates MIMIC-IV admission records and clinical notes into time-series event streams and long-context memory datasets, enabling long-horizon, multi-session interactions between agents and a clinical environment. It comprises 335 patients, with 19.72 inpatient visits per patient on average and 44.91 medical events per visit. Guided by the long-horizon decision process, we propose an evaluation taxonomy with three suites: fact-based QA, temporal reasoning, and long-horizon decision-making. This taxonomy measures how agents understand and leverage historical patient information over extended horizons. Our experiments show that while recent LLMs can make good use of explicit timestamps, they have challenges in implicit time inference; The RAG and agent memory system can improve the performance of information retrieval tasks, but the performance of decision-making tasks is highly dependent on the model's immediate context.

\keywords{Computer-Aided Diagnosis  \and Medical Agents \and  EHR.}

\end{abstract}
\section{Introduction}
In clinical practice, the diagnostic process is inherently longitudinal and time-dependent. Clinicians do not merely react to isolated symptoms; instead, they must synthesize evidence across multiple visits, diagnostic tests, and evolving treatment responses over years \cite{pmid41325597}. This capacity of \textbf{long-horizon clinical reasoning} is fundamental to high-quality care. As Large Language Models (LLMs) transition to autonomous medical agents, their ability to navigate complex trajectories in real-world Electronic Health Records (EHR) \cite{PhysioNet-mimiciv-3.1} has become the critical benchmark for clinical readiness.
\begin{table}[t]
\centering
\caption{Comparison of Medical Agent Benchmarks.}
\label{tab:benchmark_comparison}

{\fontsize{8}{9.2}\selectfont  
\setlength{\tabcolsep}{2.5pt}
\renewcommand{\arraystretch}{0.9}

\begin{tabularx}{\linewidth}{
>{\raggedright\arraybackslash}p{2.8cm}  
c c c c c
>{\raggedright\arraybackslash}X         
}
\toprule
\textbf{Benchmark}
& \textbf{Long.$^a$}
& \textbf{Ctx.$^b$}
& \textbf{EHR.$^c$}
& \textbf{Dec.$^d$}
& \textbf{Temp.$^e$}
& \textbf{Focus} \\
\midrule

MedAgentBench\cite{jiang2025medagentbench}
& $\times$ & $\times$ & $\checkmark$ & $\checkmark$ & $\times$
& EHR Tool Integration \\

AgentClinic\cite{schmidgall2025agentclinicmultimodalagentbenchmark}
& $\times$ & $\times$ & $\times$ & $\checkmark$ & $\times$
& Simulated Interaction \\

DiagBench\cite{qiu2026evolvinginteractivediagnosticagents}
& $\times$ & $\times$ & $\checkmark$ & $\checkmark$ & $\times$
& Diagnostic Trajectory \\

MedBench v4\cite{ding2025medbenchv4robustscalable}
& $\times$ & $\checkmark$ & $\checkmark$ & $\times$ & $\times$
& Medical Knowledge QA \\

ReflecTool\cite{liao2024reflectoolreflectionawaretoolaugmentedclinical}
& $\times$ & $\times$ & $\times$ & $\checkmark$ & $\times$
& Reflective Tool-Use \\

EHRSQL\cite{lee2022ehrsql}
& $\times$ & $\times$ & $\checkmark$ & $\times$ & $\times$
& Relational Querying \\

\textbf{LongMedBench}
& \textbf{$\checkmark$}
& \textbf{$\checkmark$}
& \textbf{$\checkmark$}
& \textbf{$\checkmark$}
& \textbf{$\checkmark$}
& Long-horizon Reasoning \\
\bottomrule

\multicolumn{7}{p{\dimexpr\linewidth-2\tabcolsep\relax}}{
$^a$\textit{Longitudinal}: Reasoning across multiple discrete clinical visits.\;
$^b$\textit{Context}: Whether the benchmark environment contains multi-turn context.\;
$^c$\textit{EHR}: The dataset built upon real-world EHR.\;
$^d$\textit{Decision-making}: Evaluating proactive clinical planning beyond static retrieval or SQL-based querying.\;
$^e$\textit{Temporal Sensitivity}: Evaluating understanding of clinical timeline and urgency.
}
\end{tabularx}
}
\end{table}

Although general-purpose benchmarks have emerged to evaluate long-horizon or multi-turn interactions \cite{wu2024longmemeval,shen2026tripbenchbenchmarklonghorizoninteractive,kočiský2017narrativeqareadingcomprehensionchallenge,yang2018hotpotqa,hsieh2024rulerwhatsrealcontext,ding2025medbenchv4robustscalable}, these benchmarks primarily emphasize \textbf{explicit reasoning based on information retrieval}—the ability to locate specific facts within a long context \cite{liu2023lostmiddlelanguagemodels}. They test a model's capacity to find a ``needle in a haystack", but overlook the temporal dynamics that are essential to clinical reasoning. In real-world medical scenarios, the challenge extends beyond retrieving isolated facts to understanding how a patient's state evolves over dozens of visits and how this temporal progression informs treatment strategies.

In parallel with these general limitations, current medical evaluation paradigms also fall short of this requirement, as shown in Table \ref{tab:benchmark_comparison}. Despite introducing simulated interactive environments, recent frameworks \cite{qiu2026evolvinginteractivediagnosticagents,jiang2025medagentbench,schmidgall2025agentclinicmultimodalagentbenchmark,ding2025medbenchv4robustscalable,liao2024reflectoolreflectionawaretoolaugmentedclinical,lee2022ehrsql} are \textbf{constrained by limited context windows and session counts}, emphasizing the agent's tool-call abilities and immediate prediction rather than understanding of complete clinical trajectories. Consequently, they fail to assess how models utilize extensive medical history for future clinical decisions.
 
To address these limitations, we introduce LongMedBench, a MIMIC-IV\cite{PhysioNet-mimiciv-3.1}-based benchmark that overcomes constrained context lengths and simple factual recall by curating extensive longitudinal trajectories and designing reasoning tasks that challenge agent’s temporal sensitivity. By converting 335 patient records into longitudinal event streams (averaging 19.72 visits per patient), we construct a temporally dense environment and a framework explicitly targeting long-horizon clinical reasoning. Our contributions are: (1) \textbf{A tri-level memory dataset architecture} for different granularity history evaluation. (2) \textbf{A progressive evaluation taxonomy} that spans three hierarchical tasks: \textbf{factual QA} based on timestamp or relative positioning, targeting the fact retrieval limitation of general benchmarks; \textbf{temporal reasoning} for multi-visit and event-level ordering, addressing the lack of time-sensitive evaluation in existing medical frameworks; and \textbf{long-horizon decision-making} which directly challenges the agent’s ability to navigate extensive histories and autonomously plan next-step clinical actions. (3) \textbf{Experimental findings}: we showed that while state-of-the-art LLMs can exploit explicit timestamps, they struggle with the implicit temporal reasoning required for visit-level understanding. Although RAG and memory systems improve fact retrieval performance, decision-making accuracy remains highly dependent on immediate context, highlighting a profound limitation in reasoning over long-term clinical trajectories.

\section{Methodology}

\subsection{EHR Data Processing Pipeline}
\begin{figure}
\includegraphics[width=\textwidth]{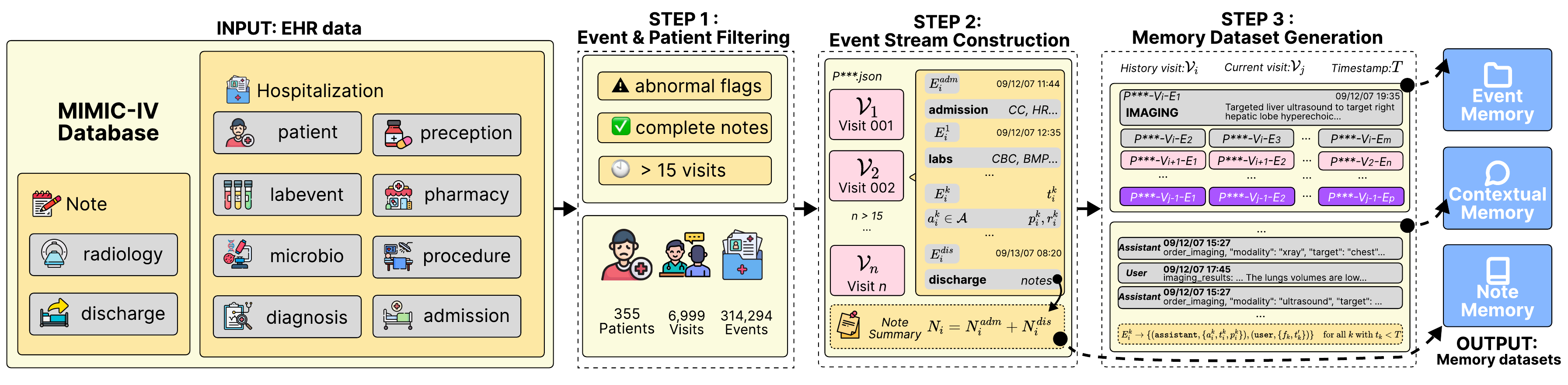}
\caption{Data processing pipeline for LongMedBench} \label{pipeline}
\end{figure}

LongMedBench is built on MIMIC-IV\cite{PhysioNet-mimiciv-3.1}, a public database that contains medical records for $>100,000$ patients. To convert static EHRs into an interactive agent environment, we design a three-stage pipeline (Figure \ref{pipeline}). 

\subsubsection{Event \& Patient Filtering} 
To reduce redundancy while preserving clinical signals, we retain only abnormal lab results and patients with complete admission/discharge records. Filtering for patients with $\ge 15$ hospitalizations yields 355 patients and 6,999 visits. With an average of 19.72 visits per patient (median=18.00, SD=5.72), this dense multi-session structure rigorously evaluates the agent's ability in cross-session memorization and temporal reasoning.




\subsubsection{Event Stream Construction}
A complete event stream will be constructed for each patient $\mathcal{P}$. The event stream $\mathcal{S}$ contains a patient's visit records $\mathcal{S}:=\{\mathcal{V}_1, \mathcal{V}_2, \dots, \mathcal{V}_n\}$, where $\mathcal{V}_i$ is the $i_\text{th}$ visit of the patient and $n\ge15$. On average, each visit $\mathcal{V}_i$ contains 44.91 medical events (median=25.00, SD=70.32).

A visit $\mathcal{V}_i$ begins with an admission event $E^{adm}_{i}$, followed by a series of specific medical events, and ends with a discharge event $E^{dis}_{i}$.
The $k_\text{th}$ medical event in $\mathcal{V}_i$ is denoted as $E^k_i=\{a^k_i,t^k_i,p^k_i,o^k_i\}$, where $a^k_i$ is an  action within a typical clinic action space $\mathcal{A}$, which includes~\textit{imaging, lab tests, medication, \emph{etc}.} $t^k_i$ is the event timestamp, and $p^k_i$ is the corresponding action argument, \textit{such as the modality (CT, X-ray, MRI) of an imaging event}. $o^k_i$ is the event result or clinical observation, \textit{like the radiology report in an imaging event}. $\mathcal{V}_i$ can therefore be expressed as a union of sequenced events: $\mathcal{V}_i = \{E^{adm}_{i}, E^{1}_{i}, E^{2}_{i},\dots,E^{dis}_{i}\}$.

In addition to the structured hospitalization data, each visit also includes two types of notes: one is radiology notes, which are parsed as imaging events; the other is discharge notes, which are summaries $N_i$ from visits $\mathcal{V}_i$, and can be parsed into admission info $N_i^{adm}$ and discharge info $N_i^{dis}$ according to logic.

\subsubsection{Memory Dataset Generation}
To evaluate how agents utilize long-context information, we design three memory modules with progressively finer granularity. From the beginning history visit $\mathcal{V}_i$, current visit $\mathcal{V}_j$ and reasoning timestamp $T$, the agent's memory access is strictly bounded to prevent event leakage.

\textbf{1. Note Memory}: $\mathcal{M}^{(i,j)}_{N}=\{N_{i},N_{i+1},\dots,N_{j-1}\}$, which contains visit-level summaries from preceding $\mathcal{V}_i$ to $\mathcal{V}_{j-1}$. The summary is high-level compressed and emphasizing global trajectory rather than detailed event recall.

\textbf{2. Event Memory}: ${\textstyle \mathcal{M}^{(i,j)}_E = \bigcup_{k=i}^{j-1}\mathcal{V}_k }$, which contains all clinical events from $\mathcal{V}_i$ to $\mathcal{V}_{j-1}$. Events are flattened into a single chronological sequence, discarding visit-level boundaries, and forming a unified patient trajectory.

\textbf{3. Contextual Memory}: $\mathcal{M}_C^{(j,T)} = \{ (r_0, m_0), (r_1, m_1), \dots, (r_P, m_P) \}$, 
where $r_k$ and $m_k$ denote the dialog role and message content. It is constructed by rewriting the event stream of current $\mathcal{V}_j$ into an LLM-style trajectory. Only events $E_j^k$ with $t_j^k<T$ are retained. 
Each $E_j^k$ is transformed into a action–feedback pair using LLM:
$\{(\texttt{assistant}, \{a^k_j,t^k_j,p^k_j\}),(\texttt{user}, \{f^k_j,{t^k_j}'\})\}$,
where $f^k_j$ and ${t^k_j}'$ denote simulated clinic feedback inferred from $o_j^k$ and corresponding timestamp. This memory simulates the instant interaction status for a medical agent.

\subsection{Benchmark Question Generation}
Based on the memory modules 
$\{\mathcal{M}^{(i,j)}_N,\mathcal{M}^{(i,j)}_E,\mathcal{M}_C^{(j,T)}\}$, 
we construct three progressively challenging task families, from \emph{factual QA}, \emph{temporal reasoning}, to 
\emph{long-horizon decision making}, reflecting increasingly global memory dependency.
\begin{figure}
\includegraphics[width=\textwidth]{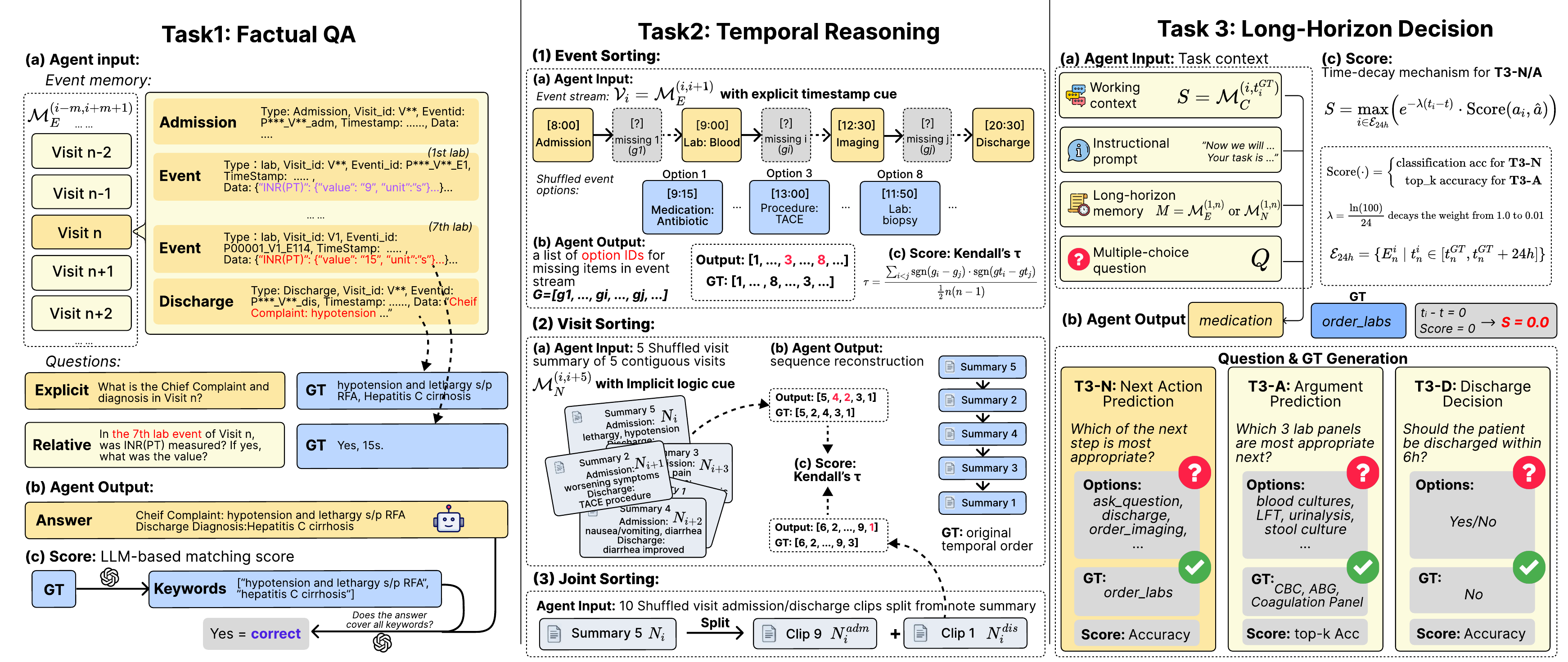}
\caption{Example of the three evaluation tasks in LongMedBench.} \label{frame3}
\end{figure}
\subsubsection{Factual QA}
This task evaluates precise retrieval and temporal alignment within the agent's event memory. From visit $\mathcal{V}_i$, a ground truth event
$E^k_i \in \mathcal{V}_i$ is randomly sampled, and is unified as questions using the following two formats: \textbf{Explicit} tasks include $t^k_i$ in the query text, evaluating direct retrieval capacity; while \textbf{Relative} questions provide only relative temporal relationships without timestamp \textit{(e.g. the medicine name in the 2nd medication event of $\mathcal{V}_i$)}.
The questions require the agent to recover full factual details $p_i^k$
from a redundant memory window $\mathcal{M}_E^{(i-m,i+m+1)}$ provided, where $m \ge0 $ is a hyperparameter to adjust the window size. We place $\mathcal{V}_i$ in the middle of the window to evaluate retrieval from non-recency-favored regions of long-context memory \cite{liu2023lostmiddlelanguagemodels}.

\subsubsection{Temporal Reasoning}
This task evaluates the agent's ability to reconstruct chronological order at different memory granularities. The evaluation score is measured by Kendall's $\tau$. Given a target event stream $\mathcal{S}$ or its visit $\mathcal{V}_i$, we design three sub-tasks using the original event streams as ground truth, each targeting at a distinct level of temporal reasoning. \textbf{Visit Cloze} tests event-level reasoning: several intervention events $E_i^k \in \mathcal{V}_i$ with timestamps preserved are masked and provided to the agent as a shuffled list. The agent must insert each event back to its original position. \textbf{Visit Sorting} evaluates visit-level ordering: five shuffled visit summaries $N\in\mathcal{M}_N^{(i,i+5)}$ must be ordered solely by clinical progression cues with their timestamps removed. \textbf{Joint Sorting} assesses integration: five visits are split into ten admission/discharge clips. The agent must correctly pair the clips from one visit and order them in time sequence.

\subsubsection{Long-Horizon Decision Making}
This task evaluates agent's next-step planning under long-term historical context, a more comprehensive evaluation.
From $\mathcal{V}_n$, we sample a ground-truth event $E_i^{GT}$, and define the agent's contextual memory as its working context $S$. The long-horizon note or event memory is also provided as a reference $M$. When generating question $Q$, we randomly select one of the following three sub-task formats:\textbf{Next Action Prediction (T3-N)} requires the agent to predict $a_i^{GT}$, \textbf{Argument Prediction (T3-A)} evaluates inference for $p_i^{GT}$, and \textbf{Discharge Decision (T3-D)} queries if the patient can be discharged within $6$h. For two \textit{prediction} tasks, as shown in Figure \ref{frame3}, a time-decay scoring mechanism is applied to recognize actions $\mathcal{E}_{24h}$, events that occur within $24h$ after $t_i^{GT}$. Unlike prior tasks, this task requires integrating immediate state $S$ with long-term memory $M$ to produce clinically coherent decisions. It evaluates true long-horizon reasoning rather than isolated memory access.

\section{Experiments and Results}
\subsection{Experiment Environments}
To verify the robustness of medical agents in long-term clinical decision, we selected representative long-context LLMs as the agent's backbone, including the closed-source frontier model \textit{gpt-5-mini}\cite{gpt52}, the lightweight one \textit{qwen-turbo}\cite{qwen3}, and the open-source high-performance one \textit{deepseek-v3.2}\cite{deepseekai2025deepseekv32}. All LLMs are deployed with default configurations.

Referring to common retrieval methods in long-horizon tasks\cite{wu2024longmemeval,maharana2024evaluating}, we set up three memory architectures: (1) \textbf{Naive Long-Context.} No external tools is used; memories form preceding visits are greedily injected into the LLM's context window. (2) \textbf{RAG.} \textit{text-embedding-3-small}\cite{text-embedding-3-small} is used to vectorize memory entries; the agent can retrieve Top-K similar memory fragments when answering the question. (3) \textbf{Agent Memory System}. A dynamically updated external storage layer is included. When answering questions, the agent can spontaneously search based on the request and recall appropriate memories. We introduce Mem0\cite{mem0}, a product-ready agent memory architecture, in the following experiments. 

\subsection{Results}
\subsubsection{Factual QA}
We evaluate \textbf{Naive Long-Context LLMs} using \textit{qwen-turbo}, comparing window sizes $\{3, 5, 7\}$ (\emph{i.e.}, $m \in \{1, 2, 3\}$ in $\mathcal{M}_E$) and full history ($m=\infty$) against the \textbf{Agent Memory system} (Mem0). Results in Table \ref{tab:factual_combined} report \textbf{Lab-T} (recall) and \textbf{Lab-F} (rejection) to assess hallucination impact. Table \ref{tab:factual_combined} reveals that \textbf{naive-LLM's performance decays severely as history grows}, dropping from 0.882 to 0.423 in explicit retrieval, with low Lab-F scores ($\le$0.570) highlighting a pervasive hallucination bias. \textbf{The performance of agent memory system is strongly correlated with the specific task type}; while Mem0 achieves near-optimal results in explicit Lab-T (0.993) and Medication (0.983), its relative performance (Overall 0.331) remains inferior. This gap reveals that: first, agents struggle to generate precise queries for imaging reports even with few-shot prompting; interestingly, when falling back to vector-based search, relative semantic queries outperform explicit ID/timestamp matching, as the latter lacks sufficient embedding density in vector space; second, the agent's memory architecture lacks a systematic indexing of relationships between events.

\begin{table}[H]
\centering
\caption{Performance on Factual Retrieval under Explicit and Relative Settings.}
\label{tab:factual_combined}

{\fontsize{8}{8.8}\selectfont
\setlength{\tabcolsep}{1pt}
\renewcommand{\arraystretch}{0.98}

\begin{tabular}{l ccccccc cccccc}
\toprule
& \multicolumn{7}{c}{\textbf{Explicit}} 
& \multicolumn{6}{c}{\textbf{Relative}} \\
\cmidrule(lr){2-8} \cmidrule(lr){9-14}

\textbf{Model}
& L-T$^f$ & L-F$^g$ & L-O$^h$
& Med. & Img. & Adm/Dis & Avg.
& L-T$^f$ & L-F$^g$ & L-O$^h$
& Med. & Img. & Avg. \\
\midrule

$m=1$
& \underline{0.96} & \textbf{0.57} & \textbf{0.83}
& \underline{0.84} & \textbf{0.89} & \underline{0.97} & \textbf{0.88}
& \textbf{0.91} & \textbf{0.58} & \textbf{0.80}
& \textbf{0.78} & \textbf{0.84} & \textbf{0.80} \\

$m=2$
& 0.92 & \underline{0.51} & 0.78
& 0.71 & \underline{0.81} & 0.92 & \underline{0.81}
& \underline{0.86} & 0.52 & \underline{0.74}
& \underline{0.64} & \underline{0.74} & \underline{0.71} \\

$m=3$
& 0.86 & 0.50 & 0.74
& 0.61 & 0.75 & 0.87 & 0.74
& 0.80 & 0.50 & 0.70
& 0.55 & 0.68 & 0.64 \\

$m=\infty$
& 0.63 & 0.29 & 0.51
& 0.28 & 0.41 & 0.46 & 0.42
& 0.52 & 0.28 & 0.44
& 0.23 & 0.36 & 0.34 \\

Mem0
& \textbf{0.99} & 0.43 & \underline{0.80}
& \textbf{0.98} & 0.07 & \textbf{0.98} & 0.74
& 0.71 & \underline{0.57} & 0.66
& 0.17 & 0.13 & 0.33 \\

\bottomrule
\multicolumn{14}{p{\dimexpr\linewidth-2\tabcolsep\relax}}{
$^f$L-T: Lab-T, recall of existing lab indicators.
$^g$L-F: Lab-F, rejection of not found lab indicators.
$^h$L-O: Overall accuracy of lab events.
}
\end{tabular}
}
\end{table}

\begin{table}[H]
\centering
\caption{Temporal reasoning performance (Kendall's $\tau$) on LongMedBench.}
\label{tab:temporal}

{\fontsize{8}{9.2}\selectfont
\setlength{\tabcolsep}{3pt}
\renewcommand{\arraystretch}{1.05}

\begin{tabular}{l ccc ccc ccc}
\toprule
& \multicolumn{3}{c}{\textit{visit\_cloze}}
& \multicolumn{3}{c}{\textit{visit\_sorting}}
& \multicolumn{3}{c}{\textit{joint\_sorting}} \\
\cmidrule(lr){2-4} \cmidrule(lr){5-7} \cmidrule(lr){8-10}

\textbf{Model}
& N & Mean & Std
& N & Mean & Std
& N & Mean & Std \\
\midrule

deepseek-v3.2
& 6776 & 0.316 & 0.352
& 765  & 0.258 & 0.523
& 2878 & 0.132 & 0.346 \\

gpt-5-mini
& 6776  & \underline{0.925} & 0.220
& 765  & \underline{0.376} & 0.552
& 2878  & \underline{0.295} & 0.439 \\

deepseek-v3.2-thinking
& 6776  & \textbf{0.969} & 0.124
& 765  & \textbf{0.424} & 0.549
& 2878  & \textbf{0.330} & 0.425 \\

qwen-turbo
& 6776 & 0.047 & 0.257
& 765  & 0.114 & 0.456
& 2878 & 0.033 & 0.256 \\

\bottomrule
\end{tabular}
}
\end{table}

\subsubsection{Temporal Reasoning}

\begin{figure}[H]
\centering

\begin{subfigure}[t]{0.48\textwidth}
    \centering
    \includegraphics[width=\linewidth]{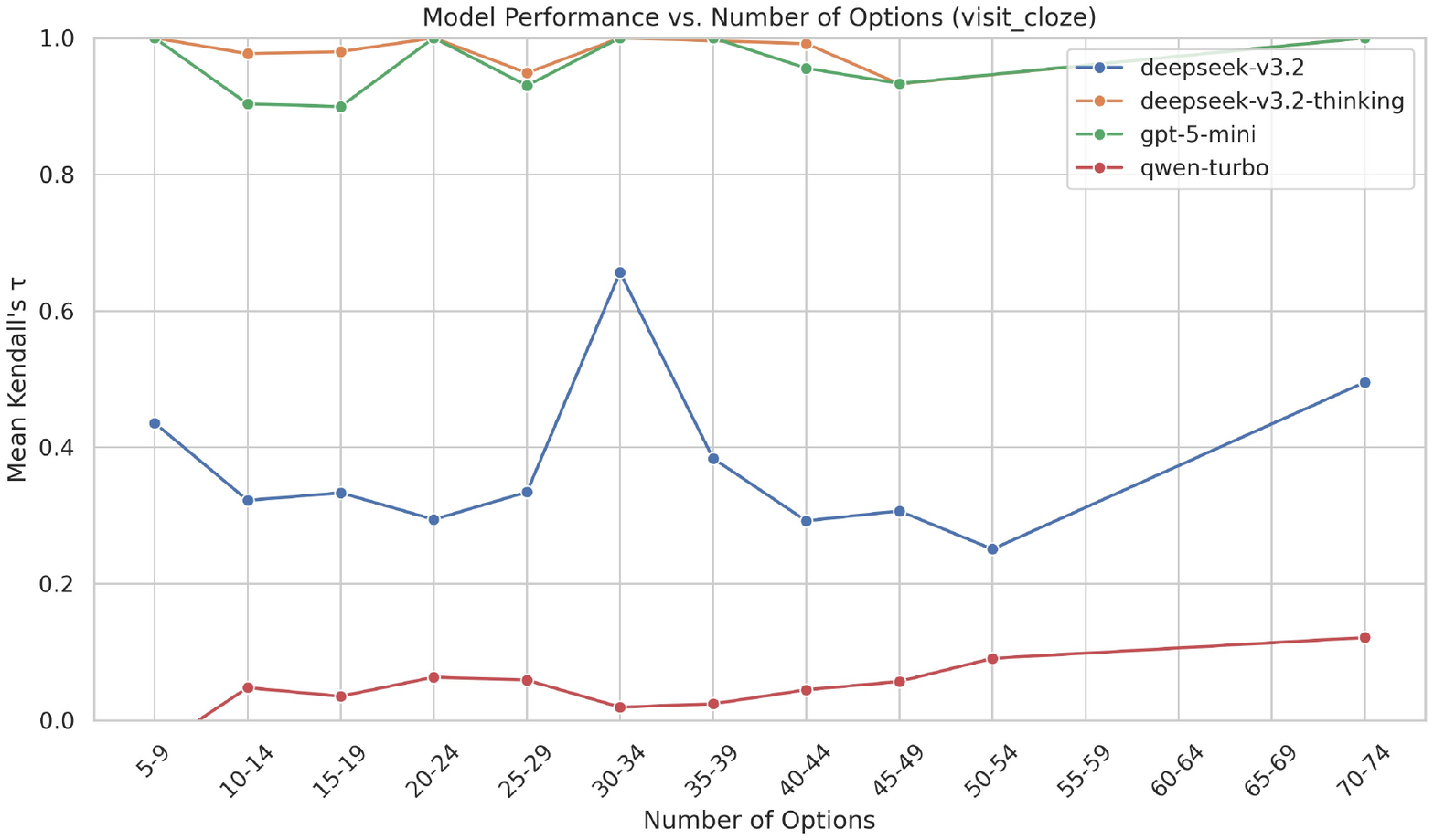}
    \caption{Mean Kendall's $\tau$ vs. number of options in \textit{visit\_cloze}.}
    \label{fig:option_complex}
\end{subfigure}
\hfill
\begin{subfigure}[t]{0.48\textwidth}
    \centering
    \includegraphics[width=\linewidth]{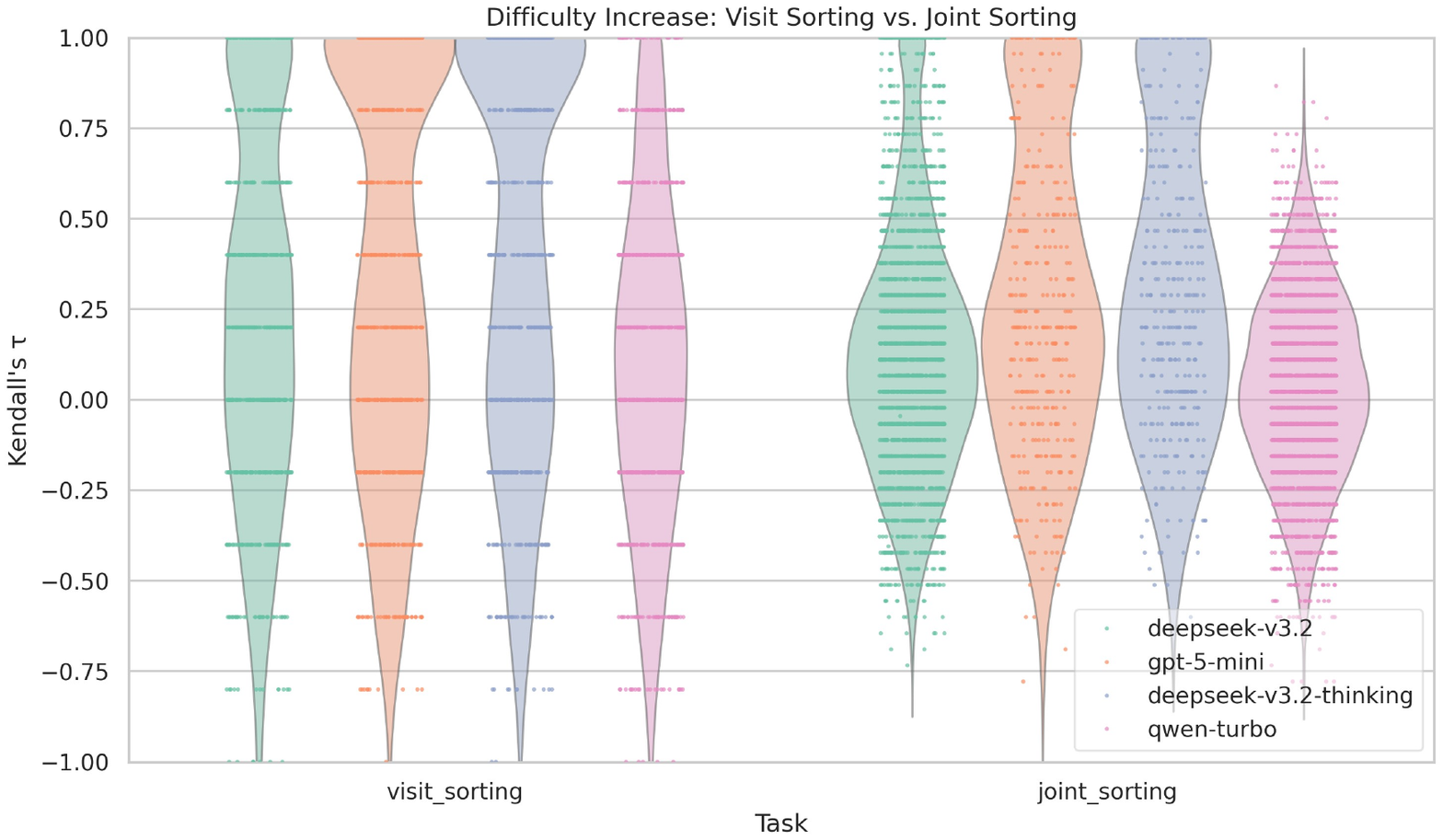}
    \caption{Comparison between \textit{joint\_sorting} and \textit{visit\_sorting}.}
    \label{fig:sorting_comparison}
\end{subfigure}

\caption{Temporal reasoning analysis.}
\label{fig:temporal_analysis}

\end{figure}

Table~\ref{tab:temporal} presents Kendall's $\tau$ for temporal reasoning across three progressive tasks. In event-level \textbf{visit cloze}, top models achieve near-perfect performance (\textit{gpt-5-mini}: 0.925), while weaker models struggle (\textit{qwen-turbo}: 0.046). Enabling thinking mode substantially boosts reasoning (\textit{deepseek-v3.2-thinking}: 0.969 vs. \textit{deepseek-v3.2}: 0.316). This confirms that \textbf{strong LLMs can effectively utilize explicit timestamp cues} for event-level ordering, regradless of the option numbers (the length of the event stream), as illustrated in Figure~\ref{fig:option_complex}.

Moving to \textbf{Visit Sorting}, where timestamps are absent so that models must sort five complete visit summaries based solely on clinical progression, performance drops sharply (best: \textit{deepseek-v3.2-thinking} 0.423). This reveals that \textbf{implicit temporal reasoning in visit-level remains challenging}.

The difficulty further escalates in \textbf{Joint Sorting}, where each visit summary is split into admission and discharge fragments, requiring simultaneous event-level pairing and visit-level sorting. Performance declines to 0.330, demonstrating the compounding complexity. Figure~\ref{fig:sorting_comparison} shows this clear descending trend, confirming that when crucial information is fragmented at both event and visit level, \textbf{implicit temporal reasoning in realistic multi-visit scenarios remains a major challenge}.

\begin{table}[H]
\centering
\caption{Long-context LLM performance on Long-Horizon Decision Making.}
\label{tab:longctx_models_compact}

{\fontsize{8}{9.2}\selectfont
\setlength{\tabcolsep}{3pt}
\renewcommand{\arraystretch}{1.05}

\begin{tabular}{l c cccc cccc}
\toprule
& & \multicolumn{4}{c}{\textbf{Note Memory}} & \multicolumn{4}{c}{\textbf{Event Memory}} \\
\cmidrule(lr){3-6}\cmidrule(lr){7-10}
\textbf{Model} & \textbf{Ctx}$^i$
& \textbf{T3-A} & \textbf{T3-D} & \textbf{T3-N} & \textbf{Avg}
& \textbf{T3-A} & \textbf{T3-D} & \textbf{T3-N} & \textbf{Avg} \\
\midrule

qwen-turbo & 128K
& 0.475 & 0.604 & \underline{0.309} & \underline{0.434}
& \underline{0.494} & 0.604 & \underline{0.309} & 0.420 \\

deepseek-v3.2 & 128K
& \textbf{0.478} & \underline{0.607} & 0.278 & 0.422
& 0.493 & \textbf{0.607} & 0.304 & \underline{0.439} \\

gpt-5-mini & 400K
& \underline{0.478} & \textbf{0.625} & \textbf{0.324} & \textbf{0.445}
& \textbf{0.494} & \textbf{0.607} & \textbf{0.335} & \textbf{0.452} \\

deepseek-v3.2-thinking & 128K
& 0.471 & 0.578 & 0.294 & 0.420
& 0.466 & 0.593 & 0.290 & 0.419 \\

\bottomrule
\multicolumn{10}{l}{$^i$\textit{Ctx}: the maximum context length of LLM.}
\end{tabular}
}
\end{table}

\subsubsection{Long-Horizon Decision Making}
We first evaluate the performance of Naive Long-Context. All models are limited to a maximum context of 128K tokens. As shown in Table~\ref{tab:longctx_models_compact}, structured event memory outperforms note memory in most models in decision making. \textit{gpt-5-mini} performs best. Notably, for \textit{deepseek-v3.2}, thinking patterns did not significantly improve decision performance. This is due to thinking mode makes the agent more conservative, attempting to ask question to get instant information rather than referring to past memories. 

To analyze the Lost-in-Middle\cite{liu2023lostmiddlelanguagemodels} effect, using \textit{qwen-turbo} as the baseline model, we (1) adjusted the number of injected historical visits and (2) implement RAG and Mem0 architecture as retrial source in separate experiments. As shown in Table~\ref{tab:agent_ablation}, although the difference is subtle, close to that under full memory conditions, increasing the context length makes the agent perform worse. The performance differences between Mem0 and RAG are minimal, both close to the baseline performance, demonstrating that existing memory augmentation architectures offer limited gains for long-term decision tasks.

\begin{table}[H]
\centering
\caption{Ablation study. Left: visit injection ($n$). 
Right: memory architectures.}
\label{tab:agent_ablation}

{\fontsize{8}{8.8}\selectfont
\setlength{\tabcolsep}{1pt}
\renewcommand{\arraystretch}{0.98}

\begin{tabular}{c l c c c c | l l c c c c}
\toprule
\multicolumn{6}{c}{\textbf{Visit Injection}} 
& \multicolumn{6}{c}{\textbf{Memory Architecture}} \\
\cmidrule(lr){1-6}
\cmidrule(lr){7-12}

$n$ & Memory & T3-A & T3-D & T3-N & Avg 
& Arch & Memory & T3-A & T3-D & T3-N & Avg \\
\midrule

0 (Base)$^j$ & -- 
& \textbf{0.49} & \textbf{0.61} & \textbf{0.32} & \textbf{0.45}
& RAG & event 
& 0.45 & 0.51 & 0.29 & 0.40 \\

2 & event 
& \underline{0.49} & \underline{0.60} & 0.32 & \underline{0.44}
& Mem0 & event 
& \underline{0.47} & \textbf{0.62} & \underline{0.31} & \textbf{0.44} \\

5 & event 
& 0.48 & 0.60 & 0.31 & 0.43
& RAG & note 
& \textbf{0.48} & \underline{0.60} & \textbf{0.31} & \underline{0.44} \\

2 & note  
& 0.48 & \underline{0.60} & \underline{0.32} & 0.44
& Mem0 & note 
& 0.46 & 0.56 & 0.29 & 0.41 \\

5 & note  
& 0.48 & \underline{0.60} & 0.32 & 0.44
& \multicolumn{6}{c}{} \\

\bottomrule
\multicolumn{12}{p{\dimexpr\linewidth-2\tabcolsep\relax}}{$^j$Baseline with no memory content injected.}
\end{tabular}
}
\end{table}

Therefore, under the current LongMedBench setting, decision quality is not strongly correlated with the amount of retrieved historical information, while \textbf{the agent performance depends more on the model’s clinical reasoning capacity within the immediate context.} Accordingly, we need to design tasks that amplify deeper reasoning requirements and higher decision interdependencies in an attempt to better reveal the limitations of current models and more effectively assess their long-term clinical reasoning capabilities. 

\section{Conclusion}
We introduce LongMedBench, a benchmark using real-world EHR data to evaluate medical agents in long-horizon clinical reasoning. By converting MIMIC-IV records into multi-session event streams, we assess agents across three memory types with varying tasks. Experiments reveal that while state-of-the-art models handle explicit timestamps well, implicit temporal reasoning remains a significant bottleneck. As history scales, models increasingly rely on immediate context rather than cross-session integration. Furthermore, while memory augmentation improves retrieval, decision-making performance is highly task-sensitive and remains limited by the model's instant reasoning capacity. Our results highlight a critical gap in handling time-dependent clinical tasks for medical agents, necessitating future optimization of cross-session memory augmentation architectures.

\bibliographystyle{splncs04}
\bibliography{ref.bib}

@misc{kočiský2017narrativeqareadingcomprehensionchallenge,
      title={The NarrativeQA Reading Comprehension Challenge}, 
      author={Tomáš Kočiský and Jonathan Schwarz and Phil Blunsom and Chris Dyer and Karl Moritz Hermann and Gábor Melis and Edward Grefenstette},
      year={2017},
      eprint={1712.07040},
      archivePrefix={arXiv},
      primaryClass={cs.CL},
      url={https://arxiv.org/abs/1712.07040}, 
}

@misc{hsieh2024rulerwhatsrealcontext,
      title={RULER: What's the Real Context Size of Your Long-Context Language Models?}, 
      author={Cheng-Ping Hsieh and Simeng Sun and Samuel Kriman and Shantanu Acharya and Dima Rekesh and Fei Jia and Yang Zhang and Boris Ginsburg},
      year={2024},
      eprint={2404.06654},
      archivePrefix={arXiv},
      primaryClass={cs.CL},
      url={https://arxiv.org/abs/2404.06654}, 
}

@inproceedings{yang2018hotpotqa,
  title={{HotpotQA}: A Dataset for Diverse, Explainable Multi-hop Question Answering},
  author={Yang, Zhilin and Qi, Peng and Zhang, Saizheng and Bengio, Yoshua and Cohen, William W. and Salakhutdinov, Ruslan and Manning, Christopher D.},
  booktitle={Conference on Empirical Methods in Natural Language Processing ({EMNLP})},
  year={2018}
}

@article{maharana2024evaluating,
  title={Evaluating very long-term conversational memory of llm agents},
  author={Maharana, Adyasha and Lee, Dong-Ho and Tulyakov, Sergey and Bansal, Mohit and Barbieri, Francesco and Fang, Yuwei},
  journal={arXiv preprint arXiv:2402.17753},
  year={2024}
}

@article{wu2024longmemeval,
      title={LongMemEval: Benchmarking Chat Assistants on Long-Term Interactive Memory}, 
      author={Di Wu and Hongwei Wang and Wenhao Yu and Yuwei Zhang and Kai-Wei Chang and Dong Yu},
      journal={https://arxiv.org/abs/2410.10813},
      year={2024} 
}

@misc{shen2026tripbenchbenchmarklonghorizoninteractive,
      title={TRIP-Bench: A Benchmark for Long-Horizon Interactive Agents in Real-World Scenarios}, 
      author={Yuanzhe Shen and Zisu Huang and Zhengyuan Wang and Muzhao Tian and Zhengkang Guo and Chenyang Zhang and Shuaiyu Zhou and Zengjie Hu and Dailin Li and Jingwen Xu and Kaimin Wang and Wenhao Liu and Tianlong Li and Fengpeng Yue and Feng Hong and Cao Liu and Ke Zeng},
      year={2026},
      eprint={2602.01675},
      archivePrefix={arXiv},
      primaryClass={cs.AI},
      url={https://arxiv.org/abs/2602.01675}, 
}

@article{lee2022ehrsql,
  title={EHRSQL: A Practical Text-to-SQL Benchmark for Electronic Health Records},
  author={Lee, Gyubok and Hwang, Hyeonji and Bae, Seongsu and Kwon, Yeonsu and Shin, Woncheol and Yang, Seongjun and Seo, Minjoon and Kim, Jong-Yeup and Choi, Edward},
  journal={Advances in Neural Information Processing Systems},
  volume={35},
  pages={15589--15601},
  year={2022}
}

@article{pmid41325597,
  title = {Knowledge-Practice Performance Gap in Clinical Large Language Models: Systematic Review of 39 Benchmarks.},
  author = {Gong, Eun Jeong and Bang, Chang Seok and Lee, Jae Jun and Baik, Gwang Ho},
  journal = {Journal of medical Internet research},
  year = {2025},
  volume = {27},
  pages = {e84120},
  doi = {10.2196/84120},
  issn = {1438-8871 (Electronic)},
  pmid = {41325597}
}

@misc{liao2024reflectoolreflectionawaretoolaugmentedclinical,
      title={ReflecTool: Towards Reflection-Aware Tool-Augmented Clinical Agents}, 
      author={Yusheng Liao and Shuyang Jiang and Yanfeng Wang and Yu Wang},
      year={2024},
      eprint={2410.17657},
      archivePrefix={arXiv},
      primaryClass={cs.CL},
      url={https://arxiv.org/abs/2410.17657}, 
}

@article{jiang2025medagentbench,
  title={MedAgentBench: A Virtual EHR Environment to Benchmark Medical LLM Agents},
  author={Jiang, Yixing and Black, Kameron C and Geng, Gloria and Park, Danny and Zou, James and Ng, Andrew Y and Chen, Jonathan H},
  journal={NEJM AI},
  pages={AIdbp2500144},
  year={2025},
  publisher={Massachusetts Medical Society}
}

@misc{ding2025medbenchv4robustscalable,
      title={MedBench v4: A Robust and Scalable Benchmark for Evaluating Chinese Medical Language Models, Multimodal Models, and Intelligent Agents}, 
      author={Jinru Ding and Lu Lu and Chao Ding and Mouxiao Bian and Jiayuan Chen and Wenrao Pang and Ruiyao Chen and Xinwei Peng and Renjie Lu and Sijie Ren and Guanxu Zhu and Xiaoqin Wu and Zhiqiang Liu and Rongzhao Zhang and Luyi Jiang and Bing Han and Yunqiu Wang and Jie Xu},
      year={2025},
      eprint={2511.14439},
      archivePrefix={arXiv},
      primaryClass={cs.CL},
      url={https://arxiv.org/abs/2511.14439}, 
}

@misc{qiu2026evolvinginteractivediagnosticagents,
      title={Evolving Interactive Diagnostic Agents in a Virtual Clinical Environment}, 
      author={Pengcheng Qiu and Chaoyi Wu and Junwei Liu and Qiaoyu Zheng and Yusheng Liao and Haowen Wang and Yun Yue and Qianrui Fan and Shuai Zhen and Jian Wang and Jinjie Gu and Yanfeng Wang and Ya Zhang and Weidi Xie},
      year={2026},
      eprint={2510.24654},
      archivePrefix={arXiv},
      primaryClass={cs.CL},
      url={https://arxiv.org/abs/2510.24654}, 
}

@article{PhysioNet-mimiciv-3.1,
  author = {Johnson, Alistair and Bulgarelli, Lucas and Pollard, Tom and Gow, Brian and Moody, Benjamin and Horng, Steven and Celi, Leo Anthony and Mark, Roger},
  title = {{MIMIC-IV}},
  journal = {{PhysioNet}},
  year = {2024},
  month = oct,
  note = {Version 3.1},
  doi = {10.13026/kpb9-mt58},
  url = {https://doi.org/10.13026/kpb9-mt58}
}

@misc{gpt52,
  author       = {OpenAI},
  title        = {GPT-5 mini Model},
  year         = {2026},
  howpublished = {\url{https://developers.openai.com/api/docs/models/gpt-5-mini}},
  note         = {Accessed 22 Feb 2026}
}

@misc{text-embedding-3-small,
  author       = {OpenAI},
  title        = {text-embedding-3-small Model},
  year         = {2026},
  howpublished = {\url{https://developers.openai.com/api/docs/models/text-embedding-3-small}},
  note         = {Accessed 22 Feb 2026}
}

@misc{qwen3,
  title        = {Qwen3: Think Deeper, Act Faster},
  author       = {QwenTeam},
  year         = {2024},
  howpublished = {\url{https://qwen.ai/blog?id=qwen3}},
  note         = {Accessed: 22 Feb 2026}
}

@misc{deepseekai2025deepseekv32,
      title={DeepSeek-V3.2: Pushing the Frontier of Open Large Language Models}, 
      author={DeepSeek-AI},
      year={2025},
}

@article{mem0,
  title={Mem0: Building Production-Ready AI Agents with Scalable Long-Term Memory},
  author={Chhikara, Prateek and Khant, Dev and Aryan, Saket and Singh, Taranjeet and Yadav, Deshraj},
  journal={arXiv preprint arXiv:2504.19413},
  year={2025}
}

@misc{liu2023lostmiddlelanguagemodels,
      title={Lost in the Middle: How Language Models Use Long Contexts}, 
      author={Nelson F. Liu and Kevin Lin and John Hewitt and Ashwin Paranjape and Michele Bevilacqua and Fabio Petroni and Percy Liang},
      year={2023},
      eprint={2307.03172},
      archivePrefix={arXiv},
      primaryClass={cs.CL},
      url={https://arxiv.org/abs/2307.03172}, 
}

@misc{schmidgall2025agentclinicmultimodalagentbenchmark,
      title={AgentClinic: a multimodal agent benchmark to evaluate AI in simulated clinical environments}, 
      author={Samuel Schmidgall and Rojin Ziaei and Carl Harris and Eduardo Reis and Jeffrey Jopling and Michael Moor},
      year={2025},
      eprint={2405.07960},
      archivePrefix={arXiv},
      primaryClass={cs.HC},
      url={https://arxiv.org/abs/2405.07960}, 
}
\end{document}